\documentclass[runningheads]{llncs}
\usepackage{caption}
\usepackage{graphicx}
\begin{document}
\title{Game-Of-Goals: Using adversarial games to achieve strategic resilience}
%
%\titlerunning{Abbreviated paper title}
% If the paper title is too long for the running head, you can set
% an abbreviated paper title here
%
\author{Asjad Khan\ \and Aditya Ghose}
\authorrunning{F. Author et al.}
% First names are abbreviated in the running head.
% If there are more than two authors, 'et al.' is used.
%
\institute{School of Computing and Information Technology, University of  Wollongong, Australia }
\maketitle              % typeset the header of the contribution
\begin{abstract}
Our objective in this paper is to develop a machinery that makes a given organizational strategic plan resilient to the actions of competitor agents (adverse environmental actions). We assume that we are given a goal tree representing strategic goals (can also be seen business requirements for a software systems) with the assumption that competitor agents are behaving in a maximally adversarial fashion(opposing actions against our sub goals or goals in general). We use game tree search methods (such as minimax) to select an optimal execution strategy(at a given point in time), such that it can maximize our chances of achieving our (high level) strategic goals. Our machinery helps us determine which path to follow(strategy selection) to achieve the best end outcome. This is done by comparing alternative execution strategies available to us via an evaluation function. Our evaluation function is based on the idea that we want to make our execution plans defensible(future-proof) by selecting execution strategies that make us least vulnerable to adversarial actions by the competitor agents. i.e we want to select an execution strategy such that its leaves minimum room(or options) for the adversary to cause impediment/damage to our business goals/plans. 

\end{abstract}
\section{Introduction}

In a rapidly evolving business landscape, organizations must make strategic decisions in adversarial environments where competitors, market forces, and regulatory bodies continuously influence outcomes. Traditional decision-making frameworks often fail to account for the dynamic and competitive nature of real-world strategic planning. This paper presents an adversarial game-theoretic approach to strategic decision-making, leveraging game-tree search methods to enhance strategic resilience and optimize decision pathways. By modeling business strategy as a two-player adversarial game, our framework enables organizations to anticipate and mitigate the impact of competitive disruptions. \cite{hassabis2016alphago,berliner1980backgammon}.

Game-theoretic models have long been employed in fields such as economics, artificial intelligence, and cybersecurity to analyze adversarial interactions. Goal models play a critical role in requirements engineering, by providing a hierarchic representation of statements of stakeholder intent, with goals higher in the hierarchy (parent goals) related to goals lower in the hierarchy (sub-goals) via AND- or OR-refinement links. Goal models encode important knowledge about feasible, available alternatives for realizing stakeholder intent represented at varying levels of abstraction. A number of prominent frameworks leverage goal models, including KAOS \cite{1319986}, i* \cite{1319986} and Tropos \cite{1319986}. In the context of business strategy, these models provide a structured mechanism for reasoning about potential counter-moves by competitors, assessing risks, and selecting optimal courses of action. Our approach is grounded in classical game theory techniques such as Minimax search with Alpha-Beta pruning and Monte Carlo Tree Search (MCTS), which enable decision-makers to explore possible future states and identify strategies that maximize long-term success while minimizing vulnerability to adversarial actions. \cite{hassabis2016alphago}. \cite{cleland2006detection}.

The core objective of this work is to develop a systematic methodology for selecting strategic execution plans that are resilient to external threats and adversarial maneuvers. We introduce a computational framework that evaluates alternative goal refinements within a hierarchical goal model, allowing businesses to dynamically adapt their strategies in response to an evolving competitive landscape. Our framework integrates state-of-the-art heuristic evaluation functions to assess the viability of different strategic options, ensuring that businesses select execution pathways that minimize exposure to risk while maximizing goal attainment.

To evaluate the effectiveness of our approach, we conduct empirical simulations that assess the performance of different game-tree search techniques in adversarial settings. We analyze how computational complexity scales with various parameters such as the number of strategic choices, branching factors, and depth of exploration. Our results demonstrate that while Minimax provides exhaustive strategic insights, it is computationally expensive, whereas MCTS offers a more scalable and efficient alternative. The findings provide key insights into how organizations can leverage game-theoretic models for real-world decision-making under uncertainty.

The remainder of this paper is structured as follows: Section 2 reviews related work on game-theoretic decision-making and strategic resilience. Section 3 details our proposed adversarial game framework, outlining its key components and theoretical underpinnings. Section 4 presents our methodology for modeling strategic decision-making as a search problem. Section 5 describes the experimental setup, and Section 6 presents empirical evaluations of different search strategies. Finally, Section 7 discusses key insights, implications, and future research directions.

\section{Decision making in adversarial settings}
Decision making in adversarial settings involves reasoning about chains of moves and counter-moves by the adversarial entities involved. A telecom provider might reason along the following lines: {\em If we offer high-speed web access at heavily discounted rates, we can rapidly build market share. Then competition can counter this by copying our strategy of heavily discounted data offers. We can counter their move by sustaining the heavy discounting for a longer period of time by raising funds via an IPO where our dominant market share will work in our favour. But the competition could counter this by building a larger war chest of funds via mergers and acquisitions....} This is precisely the form of reasoning that would be used by a chess player or a Go player. A two-player adversarial game of perfect information might therefore be a useful abstract lens through which we might understand business decision making. Viewing this as a 2-player game is a convenient abstraction. While a real business decision-making setting consists of multiple players, including the business whose decisions we seek to support, its various competitors, its suppliers and partners as well as the government, industry and regulatory bodies. We simplify this picture to one consisting of the business and the rest of the world (i.e., the {\em environment} actor). By abstracting thus, we model all possible adversarial moves by competitors, or moves which ultimately have potentially negative effects on the business by its suppliers, by the government and by regulatory agencies as actions taken by the {\em environment} agent. Assuming that this is a game of perfect information is also generally valid, since all moves taken by other actors that negatively impact the business (or their effects) will ultimately be visible to the business (similarly, actions taken by the business will be visible to the other actors). The turn-taking aspect of 2-player games is also a convenient abstraction. While actual business actions do not necessarily involve turn-taking, much useful analysis can be performed by assuming this is the case (and avoiding more complex models).

The decision problem we seek to support is that of deciding which of multiple competing goal refinements a business should seek to implement. In general, goal models can offer multiple alternative refinements (OR-refinements) and the choice of refinement can sometimes represent a make-or-break decision for a business. We model this decision as a choice of a move in the class of games discussed above. 

We reason with game trees, where each node is a move made by a given player. Each node in a game tree is ultimately labelled with a {\em payoff} value. We view each player as either the {\em maximizing player} (who seeks a higher payoff value) or the {\em minimizing player} (who seeks a lower payoff value). Thus a winning state for the maximizing player in a game such as chess is labelled with a ``+1" while a winning state for the minimizing player is labelled with a ``-1" (draws are labelled with a``0"). Each alternating level represents possible moves by either the maximizing or minimizing player. We bring to bear game-tree search techniques to obtain labels for each of the move choices facing the business (we might simplify matters by assuming that the business is always the maximizing player. The specific game-tree search techniques we explore in this paper are {\em minimax search with $\alpha-\beta$ cutoffs} and {\em Monte Carlo Tree Search}. Minimax search is the best-known game-tree search technique, where the idea is to explore the game tree upto a certain depth, determined by resource bounds, such as the time allowed to a given chess player to decide a move (ideally, we would want to explore the tree up to the leaf/end-game nodes but for most hard games such as chess but this typically exceeds available time/resource bounds). Instead, we explore the tree to some {\em cutoff} depth at which we view the nodes as {\em pseudo-leaf} nodes and try to guess the ``goodness'' of these nodes using a heuristic {\em evaluation function}. 

Once we have labels on the pseudo-leaf nodes, we propagate these labels up the tree using the following rules: If a node represents a maximizing player's move, the label of the child node with the highest payoff value is selected as the label of the parent node. If a node corresponds to a minimizing player's move, the label of the child node with lowest payoff value is selected as the label of the parent node. At the final step, we obtain a label on all of the nodes available as options (i.e., actions that could be taken by the business, which we have chosen to view as the maximizing player). The business then selects that action that is labelled with the highest payoff value. This choice represents the most resilient course of action available to the business given what is known about the capabilities of the adversarial {\em environment} agent, and given the extent to which the game tree could be searched under applicable resource bounds (given by the cutoff depth).

$\alpha$ and $\beta$ cutoffs leverage upper and lower bounds to prune the game tree. Minimax search with $\alpha$-$\beta$ cutoffs has been effective in defeating world champion players in the games of checkers \cite{berliner1980backgammon} and chess \cite{DEEPBLUE}.

An alternative approach to game tree search (of particular interest because of its recent success with the Go playing system called AlphaGo \cite{hassabis2016alphago}) is Monte Carlo Tree Search (MCTS). In MCTS, we replace the application of an evaluation function on a pseudo-leaf node with Monte Carlo sampling. This involves executing a certain number of random ``playouts'' or {\em simulations} from each pseudo-leaf node where a playout involves exploring the full game tree rooted at that node up to the end-game or leaf nodes, which then enables us to obtain a payoff label for the node of interest using the same principles as minimax search discussed above. \cite{hassabis2016alphago}.
\section{Reasoning with augmented game trees}
The problem we aim to solve is that of helping organizations select amongst alternative goal refinements (OR-refinements). Given a goal model that delimits that space of goals and subgoals that an organization can seek to satisfy, this is a critical (and indeed, only) decision problem to be solved. An AND-refinement of a goal is a statement of {\em know-how} that tells the organization how to achieve a parent goal (although without sequencing information, and thus falling short of being a full procedure or process model). OR-refinements offer alternative specifications of know-how for a given parent goal. We model the problem by viewing {\em actions} that take us closer to goal or subgoal satisfaction as game moves. The available goal model for the business is therefore not the game tree, but the goal model serves to constrain a different data structure, which we will henceforth call an {\em augmented game tree} (and which we will discuss in detail below).
We observe the following:
\begin{itemize}
\item Our aim is to explore the consequences of selecting a given OR-refined subgoal to pursue.
\item There are potentially multiple such OR-refined sub-goals for any goal in a goal tree (and multiple alternative sub-subgoals for a given subgoal, and so on).
\item In addition to having access to the goal model that lays out the available OR-refinements, we also know the current state of the business environment.
%\item To understand how robust a given choice of subgoal refinement might be, we explore a game tree that consists of states and actions, rooted in a state that satisfies the given subgoal.
\item Suppose we are in a state $s$ where we have a choice between decomposing a given goal $G$ into one of two OR-refined subgoals: $g_1$ and $g_2$. To decide if $g_1$ is the preferred decomposition, we explore the game tree rooted in $s$ where $g_1$ serves as the ``win” condition. In other words, we prefer states of the world where we are maximally proximal to a $g_1$-satisfying state. The application of a game-tree search technique, as discussed above, eventually provides us with a payoff label on our current state. When we have payoff labels on the current state obtained by considering all of the available choices (in this case $g_1$ and $g_2$), we pick that choice that gives us the highest payoff value. Note that this is slightly different from what is done in traditional game tree search, where we use labels on the available moves. Here, choosing to pursue one subgoal over another is a move only in an abstract sense, but not in the concrete sense of a specific action to take. Instead, the payoff label associated with a given subgoal provides an indication of the extent to which the corresponding subgoal can be satisfied (and maintained, given the depth of the tree being searched), given the possible adversarial actions of the environment agent. The payoff label associated with a given subgoal is thus an indicator of the {\em robustness} or {\em resilience} of any strategy that seeks to satisfy the overall organizational goal(s) by selecting that subgoal as the appropriate OR-refinement.
\item The idea of maintaining goal satisfaction, as discussed above, deserves special attention. Game trees in the traditional sense are {\em bounded} (i.e., we ultimately achieve an end-game state). The game of business decision-making, on the other hand, is {\em unbounded} since there is generally no clear end-game state.  A business generally operate on the assumption that there some specific point in the future when it will cease to operate (except where stakeholders have clear exit strategies and such). Search over the game tree is therefore bounded by a {\em cutoff} determined by resource bounds. This is consistent with what is done in minimax search, but we apply this also to MTCS. In other words, we are not content to stop search once a goal-satisfying state is reached since the adversarial actions of the environment player can prevent the {\em maintenance} of that goal-satisfying state of affairs. Thus, a telecom provider might achieve a state of market dominance but then be unseated from that position by the actions of an adversarial player.
\end{itemize}
We make the following assumptions. These are intended to make our proposal quite general and with widespread applicability.
\begin{itemize}
\item The set of available actions to both the business and the enviroment agent is context-dependent and dynamically generated. 
\item Actions are represented by their postconditions. Thus, if two actions generate the same effect, they will be viewed as being the same action for our purposes.
\item Associated with the notion of an action is the noton of a {\em state}. The current state of the environment determines what both the business and the environment agents can do. The act of executing an action in a given state leads to {\em state update}. We assume that we have access to a state update operator such as the Possible Worlds Approach (PWA) or the Possible Models Approach (PMA). In general, state update leads to possibly many non-deterministic outcomes. In other words, if we are in a given state and execute an action (as specified by its postconditions), the updated state could be any one of the set of outcome states. Design-time analysis will not tell us what the specific outcome state will be (we will only find out at runtime).

We assume that we have a background knowledge base $KB$ that encodes domain constraints, business policies, compliance rules etc., and that actions are represented by their postconditions. In this paper, we use the PWA state update operator (but that is entirely a matter of convenience - hence PWA could be replaced with another state update operator while leaving the substance of our proposal unchanged). The application of the PWA state update operator $\oplus_{KB}$ on state $s$ with action $a$, given the background knowledge base $K$ is given by:\\
$s\oplus_{KB} a = \{a \cup s'\mid s\subseteq s', s'\wedge a\wedge KB \not\models\bot$ and there does not exist any $s''$ such that $s \subset s''\subseteq s$ and $s''\wedge a \wedge KB \not\models \bot\}$\\
Intuitively what this operator does is commit to the success of action $a$ (hence $a$ appears in all result states) while changing the current state minimally to resolve inconsistencies caused by the action. For instance, if the light is on in a room in the initial state, and an action which has the effect of turning the light off is executed, we end up with an inconsistency (the current state suggests that the light is on while the action effects suggest that the light is off). We resolve the inconsistency by retaining the effects of the action and as much of the description of the prior state as is consistent with the action effects (hence we discard the proposition that the light is on, but all other parts of the prior state description - these are not impacted by the action - are carried over). Since inconsistencies can be resolved, in the general case, in multiple different ways, we face the prospect of multiple non-deterministic outcomes.

%\item We leverage a notion of {\em goal model conformance} for actions. To define this, we need to define a notion of a {\em maximally refined goal} in a goal model satisifed by a given state. A goal $g$ in a goal model $GM$ a maximally refined goal associated with a state $s$ if and only if the following conditions hold:
\end{itemize}
{\bf Capability models:} The capabilities of the {\em business} and {\em environment} players are represented as {\em condition-action} rules. If a {\em condition} (or pre-condition) holds in a given state, then the corresponding action can be executed in that state (resulting in the post-conditions of the action becoming true). Each player has a {\em capability set} where each capability is a condition-action rule. In the case of the organization (or business) player, the goal model of the organization constrains its capability set (and vice versa) as we shall discuss below. In the case of the adversarial {\em environment} player, we assume that the sum total of our knowledge about the behaviour of this player is encoded in its set of condition-action rules (we acknowledge that this assumption can be violated - we sometimes know more, and sometime less, about the environment players capabilities). These capabilities can be learnt from a history of interactions (but we leave this for future work).\\
{\bf Augmented game trees:}
Fig. 1 illustrates an augmented game tree. Traditionally, all nodes in a game tree represent game moves. In our setting, an augmented game tree contains alternating levels consisting of nodes of two types: {\em sate nodes} and {\em action nodes}. Such trees are always rooted in a state node. A state node determines what actions can be taken by the player whose turn it is to make a move (these are precisely those actions for which the condition in the corresponding condition-action rule is satisfied in the state node in question). There can thus be potentially many child action nodes for a parent state node. There can also be many child state nodes for a given parent action node. This is because state update caused by executing an action in a given state can lead to potentially many non-deterministic outcome states (as discussed above). We use the following procedure for the bottom-up propagation of payoff labels. Conservatively anticipating the worst case, the label for an action node is the lowest of the payoff labels for the child state nodes of a given action node (in other words, we assume that the state that actually accrues from the application of the state update operator is the worst from the perspective of the maximizing player. If it is the minimizing player's turn to select an action in a given state, it will pick that action that is labelled with the lowest payoff value. If, on the other hand, it is the maximizing player's turn to select an action in a given state, it will select the action with the highest payoff value. Augmented game trees as presented here are similar to trees for stochastic games such as backgammon \cite{berliner1980backgammon} since the uncertainty associated with the outcome of state update parallels the uncertainty associated with the roll of dice. But they also differ, in particular in the manner in which states are generated.  
\\
{\bf The evaluation function and the role of non-functional objectives:} The evaluation function (used both for MINIMAS search and MCTS) asseses the ``goodness" of a given state. A simple evaluation function (and one that we have used in the implementation reported in this paper) assesses the proximity of a state to the current goal we need to deliberate on (i.e., the subgoal $g$ being considered as a candidate refinement - the decision problem being one of selecting between $g$ and the other alternative OR-refinements of the parent goal). In concrete terms we project the current state onto the vocabulary of the goal assertion and then compute the Hamming distance (e.g., if the goal assertion is $p\wedge q\wedge r$ and the projection of the current state onto the goal vocabulary of $p, q, r$ gives us $p, \neg q, \neg r$, then we have a Hamming distance of 2. We only cutoff search at a level consisting of state nodes, hence our pseudo-leaf nodes will always be state nodes on which this evaluation function can always be applied. We can also create evaluation functions that represent weighted sums of a proximity measure to the goal, plus a variety of non-function performance measures.
\\
{\bf How the goal model constrains actions:} We might think of goals in a goal model as representing the {\em capabilities} of the business. Formally speaking, a given goal can be refined in potentially infinitely many ways. There are infinitely many sentences that are strictly stronger than the sentence representing a parent goal (and thus entail the parent goal). Each of these sentences is therefore a valid OR-refinement of the parent goal. There are similarly infinitely many ways in which a parent goal could be AND-refined (satisfying the constraints proposed by van Lamsweerde et al \cite{LAMSWEERDE}). Organizations do not possess an infinite range of capabilities, hence only the set of {\em feasible capabilities} (i.e., the set of conditions that an organization is capable of bringing aboout) appear in an organization's goal model. Some accounts of goals provide for pre- and post-conditions of goals. A condition-action rule can thus be viewed as a goal in a goal model. One option for us is to take the position that the only actions available to the business are those that appear as goals in the business' goal model. In other settings, we might have independent access to the full range of actions/capabilities in the business. In this case, the {\em capability set} and the {\em goal model} would constrain each other in the following ways:
\begin{itemize}
\item {\em Capability completeness:} Here we check whether the {\em capability set} is able to achieve all of the goals in the goal model. Formally, for every goal/subgoal $g$ in the goal model, there must exist some subset $\{c_1, c2, \ldots, c_n\}$ of the capability set $C$ such that there exists at least one state $s$ in $(c_1 \oplus_{KB} c_2)\oplus_{KB} c_3)\oplus_{KB} \ldots \oplus_{KB} c_n\}$ such that $s \models g$. A stricter version of this requirement requires that this property hold for all states $s$ obtained by executing that sequence of capabilities. Note that we have not specified what the initial state should be from which the capability sequence is executed. Here too we can require that there exist at least one initial state $s_0$ satisfying the property above. We could also require this of all initial states but this would sometimes be unrealistic (such as requiring that a shipping company achieve market dominance as a telecom provider in the space of one fiscal year).
\item {\em Goal completeness:} We can similarly require that the goal model reflect all of the conditions that a business is capable of bringing about. We omit the formalization here for brevity.
\end{itemize}

%\begin{itemize}
%\item $s\models g$
%\item There is no successor node $g'$ to $g$ in $GM$ such that $s\models g'$
%\end{itemize}
%We will describe an action pair $\langle a, a'\rangle$ as being {\em goal model conformant} relative to a goal model $GM$ if and only if:
%\begin{itemize}
%\item $a$ and $a'$ are {\em contiguous actions} (i.e., $a$ is executed immediately before $a'$)
%\item There exists
%\end{itemize}
%An action sequence $\langle a_1,a_2, \ldots, a_k\rangle$ rooted in states $s$ and goal $g$ will be deemed to be {\em goal model conformant} with respect to goal model $GM$ if and only if:
%\begin{itemize}
%\item $g$ is a goal described in $GM$
%\item $s\models g$
%\item Every state $s'$ that is a non-deterministic outcome of the execution of $a_1$ is $s$ satisfies the property that $s'\models g'$ for some child goal $g'$ of $g$ in $GM$
%\item For every contiguous action pair $\langle a_i, a_{i+1}\rangle $
%\end{itemize}

%state  We shall view an action executed in a state $s$ as being {\em conformant} with a 

 \begin{figure}[h!]
       \centering
      \includegraphics[scale=0.5]{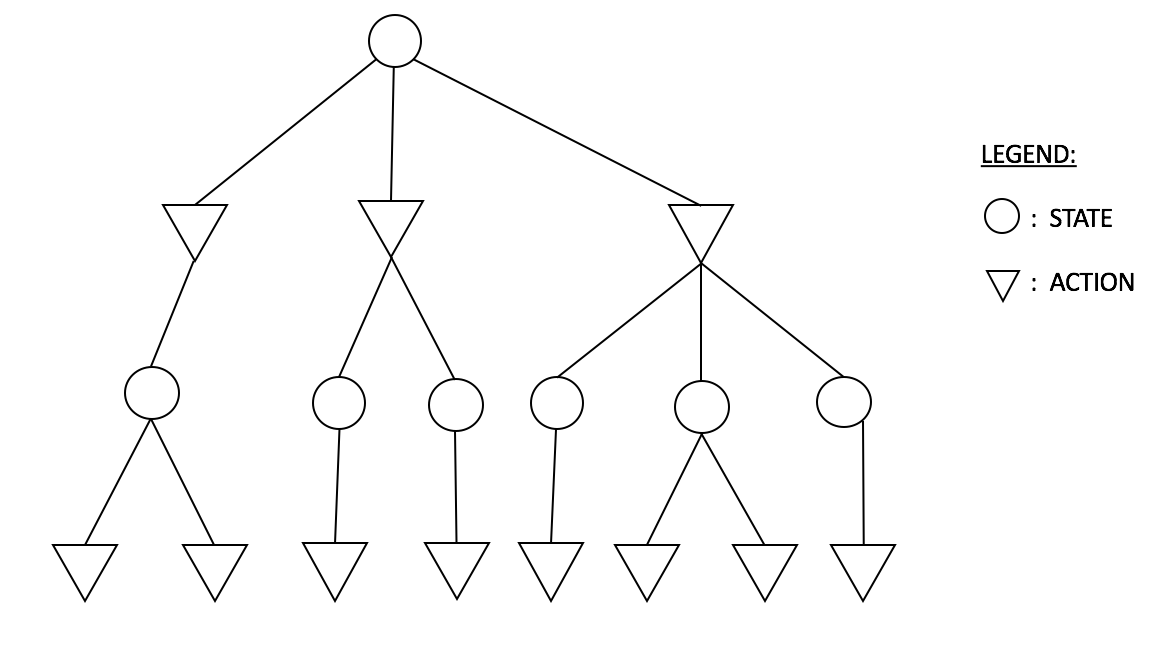}
        \caption{Game Play}
   \end{figure}
   
%\section{Adversarial analysis with BIM models}

\section{Empirical Evaluation}

  Our empirical evaluation investigates the feasibility of the proposed adversarial game-based strategic decision-making framework. Specifically, we examine how decision time scales with the number of propositional variables, the branching factor determined by condition-action rules, and the depth of game-tree exploration.

The experiment consists of a simulation similar to a two-player game where the environment (adversary) and business each take turns under the assumption of a maximally adversarial model of the environment. We assume that a given business can take certain actions (having one or more effects) to achieve a certain sub-goal. These sub-goals are represented using a set of propositional variables. We start with an initial state where these variables have been assigned random True or False values, representing the current state of the business. The actions available to a business are determined by the condition-action rules. While the condition-action rule set can change dynamically during execution, for simplicity, we assume a static set. We iterate through the condition-action rule database to check whether the current state entails any conditions specified in the rule set. If one or more rules are applicable at the current state, actions can be taken to ensure the effects defined in the applicable rules become true. This step involves using an efficient SAT-solver, which takes as input a set of clauses and determines whether they are satisfiable. After the business has made its move, the environment makes its move, simulated by impeding the achievement of certain sub-goals. Once an action is executed, the state is updated using a state update operator function (as defined in Section 3), which incorporates the effect of the action, the current state, and the knowledge base. We use a random K-CNF Generator tool to generate a set of uniform random 3-SAT instances, ensuring that clauses are unique and do not contain the same variable twice. We consider cases where the number of variables is 60 and 65, generating clauses with a clause-to-variable ratio drawn from the interval [3.5-5.5]. To determine the next state, we compute similarity using Hamming distance and select states most similar to the current one.

  \subsection {Effects of Depth Cut-Off on Performance}

A key parameter in game-tree search is the search depth cut-off, which determines how far ahead the algorithm explores possible moves. Since the game is theoretically unbounded, a reasonable cut-off must be chosen based on computational resources. As seen in Figure 1, increasing the depth cut-off causes an exponential rise in compute time for both Minimax with Alpha-Beta Pruning and MCTS. Minimax is particularly affected due to its exhaustive search strategy, while MCTS scales more efficiently but still shows increasing compute costs.

\setlength{\belowcaptionskip}{-10pt}

% First figure independently
\begin{figure}[h!]
    \centering
    \includegraphics[width=0.7\textwidth]{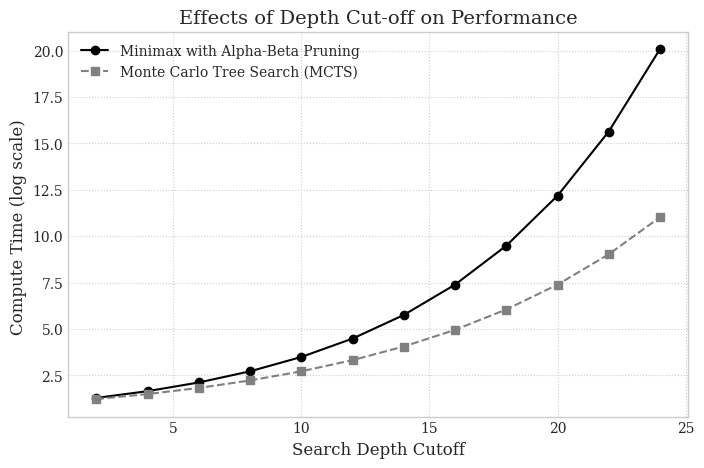}
    \caption{\small Effects of Depth Cut-off on Performance}
    \label{fig:depth_cutoff}
\end{figure}

\subsection {Effects of Simulation Count on MCTS Performance}

Unlike Minimax, MCTS relies on randomized simulations to estimate the best move. Figure 2 demonstrates the impact of increasing the number of simulations on compute time. The relationship follows a logarithmic trend, where compute time grows significantly at first but gradually plateaus, suggesting diminishing returns beyond a certain number of simulations. This suggests that practitioners using MCTS in real-world strategic planning should tune the number of simulations based on available computational resources to balance accuracy and efficiency.

\subsection { Comparative Performance of Search Algorithms}

We compared three search strategies: Minimax with Alpha-Beta Pruning, which is a deterministic approach with exponential complexity; Monte Carlo Tree Search (MCTS), a sampling-based approach that scales more efficiently; and the Stochastic Game Model (Dice Roll), a probabilistic approach used in stochastic games. As shown in Figure 3, Minimax exhibits the highest compute time, followed by MCTS, with the stochastic model performing best. The reason is that Minimax explores all possible outcomes, whereas MCTS samples only the most promising branches. The stochastic model, while computationally efficient, may not always find the best strategy due to its probabilistic nature. \cite{hassabis2016alphago}.

% First figure independently
\begin{figure}[h!]
    \centering
    \includegraphics[width=0.7\textwidth]{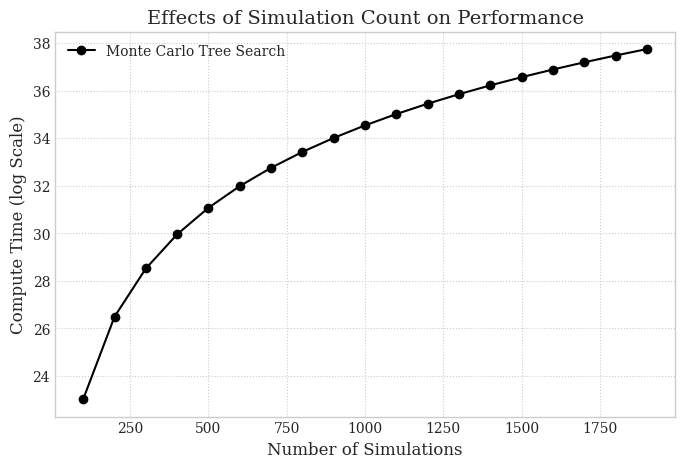}
    \caption{\small Effects of Depth Cut-off on Performance}
    \label{fig:depth_cutoff}
\end{figure}

% First figure independently
\begin{figure}[h!]
    \centering
    \includegraphics[width=0.7\textwidth]{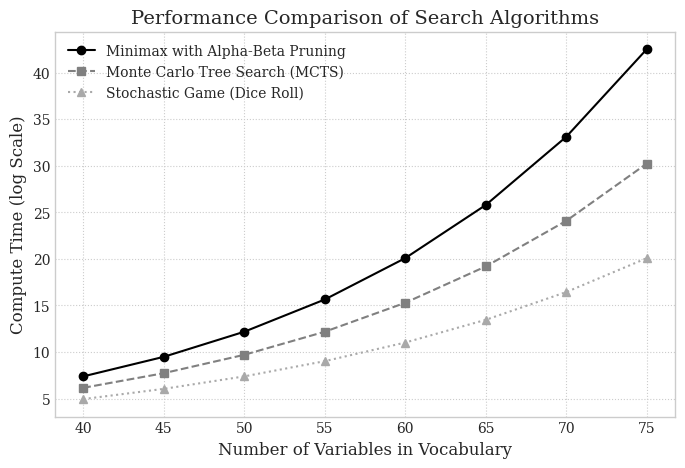}
    \caption{\small Effects of Depth Cut-off on Performance}
    \label{fig:depth_cutoff}
\end{figure}

%\subsection{Simple minimax search with $\alpha-\beta$ cutoffs}

%\subsection{Simple MCTS}

%\subsection{$alpha$-$beta$ cutoff search in settings with non-deterministic action outcomes}
Minimax is computationally expensive and impractical for deep searches, making it unsuitable for real-time decision-making in rapidly changing business environments. MCTS provides a scalable alternative by focusing on the most promising moves rather than exhaustive search, making it a better choice for adaptive strategic planning. Stochastic models trade accuracy for efficiency, making them useful when computational resources are severely constrained. Business strategy modeling must balance depth of search, accuracy, and compute time, with MCTS emerging as the best overall choice. These insights suggest that businesses can leverage adversarial game models with carefully tuned parameters to improve strategic resilience while remaining computationally feasible.

\subsection{Comparative Performance of Search Algorithms}

We compared three search strategies: Minimax with Alpha-Beta Pruning, which is a deterministic approach with exponential complexity; Monte Carlo Tree Search (MCTS), a sampling-based approach that scales more efficiently; and the Stochastic Game Model (Dice Roll), a probabilistic approach used in stochastic games. As shown in Figure 3, Minimax exhibits the highest compute time, followed by MCTS, with the stochastic model performing best. The reason is that Minimax explores all possible outcomes, whereas MCTS samples only the most promising branches. The stochastic model, while computationally efficient, may not always find the best strategy due to its probabilistic nature. \cite{hassabis2016alphago}.

\section{Related work}
Game-theoretic approaches to decision-making have been extensively studied in various domains, including artificial intelligence, economics, and cybersecurity. Classical works in game theory, such as those by von Neumann and Morgenstern, laid the foundation for strategic reasoning in adversarial settings \cite{berliner1980backgammon}. More recent advances have focused on leveraging machine learning to enhance decision-making in competitive environments \cite{hassabis2016alphago,mikolov2013efficient}.

In the field of artificial intelligence, Minimax search with Alpha-Beta pruning has been widely used in strategic decision-making, particularly in board games such as chess and Go \cite{berliner1980backgammon,hassabis2016alphago}. Monte Carlo Tree Search (MCTS) has emerged as a powerful alternative, demonstrating remarkable success in complex, high-dimensional decision spaces \cite{hassabis2016alphago}.

From a business and economics perspective, adversarial game models have been applied to strategic planning and market competition. Researchers have explored how organizations can anticipate competitor actions and optimize decision-making through predictive modeling \cite{boella2008reasoning,allen1983maintaining}. Process mining techniques have also been utilized to extract actionable insights from business process data \cite{ProM,HSRS}.

The role of requirements engineering in adversarial decision-making has also been explored. Studies have investigated how non-functional requirements, such as security and compliance, influence strategy formulation \cite{cleland2006detection,JDDS,GSSD}. Additionally, the use of data-driven approaches to inform business strategies has gained prominence, with studies emphasizing the importance of mining customer requirements and analyzing behavioral patterns \cite{QZJZ,SCJM,GP,KDFR}.

Multi-agent systems and reasoning about norms have been key areas of research in adversarial settings, particularly in regulatory compliance and institutional decision-making \cite{blevi2015discovery,teinemaa2015diagnostics}. The study of sequential pattern mining has also provided insights into identifying strategic trends over time \cite{FZLC,SG}.

In summary, while significant research has been conducted in game theory, machine learning, and strategic decision-making, our work uniquely integrates these perspectives into a unified computational framework for adversarial decision-making in business strategy. By leveraging both classical and modern game-theoretic techniques, we aim to enhance the resilience of strategic execution plans in competitive environments.

The next section presents our proposed framework, detailing its theoretical underpinnings and methodological approach.
\section{Conclusion}

This paper presented a novel approach to strategic decision-making in adversarial business environments using adversarial game-tree search methods. By modeling business strategies as two-player adversarial games, we demonstrated how decision-makers can anticipate and counteract adversarial moves effectively. Through empirical evaluation, we showed that Minimax with Alpha-Beta Pruning, Monte Carlo Tree Search, and stochastic game approaches exhibit distinct computational and strategic trade-offs. Our findings indicate that MCTS provides a well-balanced approach, offering scalability and computational feasibility while maintaining robust decision-making capabilities. Minimax, although exhaustive, is computationally expensive and unsuitable for real-time applications, whereas stochastic models provide efficiency at the cost of accuracy. These results highlight the importance of selecting appropriate search algorithms based on the strategic needs and computational constraints of a given business context. Future work will focus on extending this framework by incorporating machine learning techniques to refine strategy prediction and improve decision-making efficiency in complex, multi-agent environments. \cite{mikolov2013efficient}.

%
% ---- Bibliography ----
%
% BibTeX users should specify bibliography style 'splncs04'.
% References will then be sorted and formatted in the correct style.
%
\bibliographystyle{splncs04}
\bibliography{main}

\end{document}